\documentclass{article}

\usepackage{PRIMEarxiv}

\usepackage{multirow}
\usepackage{amssymb}
\usepackage{amsmath}
\usepackage{xcolor}
\usepackage{hyperref}

\usepackage[utf8]{inputenc} 
\usepackage[T1]{fontenc}    
\usepackage{hyperref}       
\usepackage{url}            
\usepackage{booktabs}       
\usepackage{amsfonts}       
\usepackage{nicefrac}       
\usepackage{microtype}      
\usepackage{lipsum}
\usepackage{fancyhdr}       
\usepackage{graphicx}       

\usepackage{enumitem}
\usepackage{graphicx}
\usepackage{subcaption}
\usepackage{url}
\usepackage{xcolor}
\usepackage{adjustbox}

\usepackage{amssymb}
\usepackage{amsmath}

\usepackage{placeins}
\usepackage[ruled]{algorithm2e}
\SetKwSty{textbf}
\SetKwComment{Comment}{/* }{ */}

\usepackage{booktabs}
\usepackage{multirow}
\usepackage{pifont}
\usepackage{multicol}

\title{Replay Consolidation with Label Propagation for Continual Object Detection}

\author{
  Riccardo De Monte \\
  University of Padova \\
  Padova, Italy \\
  \texttt{riccardo.demonte@phd.unipd.it} \\
   \And
  Davide Dalle Pezze \\
  University of Padova \\
  Padova, Italy \\
  \texttt{davide.dallepezze@unipd.it} \\
   \And
  Marina Ceccon \\
  University of Padova \\
  Padova, Italy \\
  \texttt{marina.ceccon@phd.unipd.it} \\
   \And
  Francesco Pasti \\
  University of Padova \\
  Padova, Italy \\
  \texttt{francesco.pasti@dei.unipd.it} \\
   \And
  Francesco Paissan \\
  Fondazione Bruno Kessler \\
  Trento, Italy \\
  \texttt{fpaissan@fbk.it} \\
   \And
  Elisabetta Farella \\
  Fondazione Bruno Kessler \\
  Trento, Italy \\
  \texttt{efarella@fbk.it} \\
   \And
  Gian Antonio Susto \\
  University of Padova \\
  Padova, Italy \\
  \texttt{gianantonio.susto@unipd.it} \\
   \And
  Nicola Bellotto \\
  University of Padova \\
  Padova, Italy \\
  \texttt{nicola.bellotto@unipd.it} \\
}

\begin{document}
\maketitle

\begin{abstract}
Continual Learning (CL) aims to learn new data while remembering previously acquired knowledge.
In contrast to CL for image classification, CL for Object Detection faces additional challenges such as the missing annotations problem.
In this scenario, images from previous tasks may contain instances of unknown classes that could reappear as labeled in future tasks, leading to task interference in replay-based approaches.
Consequently, most approaches in the literature have focused on distillation-based techniques, which are effective when there is a significant class overlap between tasks.
In our work, we propose an alternative to distillation-based approaches with a novel approach called Replay Consolidation with Label Propagation for Object Detection (RCLPOD).
RCLPOD enhances the replay memory by improving the quality of the stored samples through a technique that promotes class balance while also improving the quality of the ground truth associated with these samples through a technique called label propagation.
RCLPOD outperforms existing techniques on well-established benchmarks such as VOC and COC.
Moreover, our approach is developed to work with modern architectures like YOLOv8, making it suitable for dynamic, real-world applications such as autonomous driving and robotics, where continuous learning and resource efficiency are essential.
\end{abstract}

\keywords{Continual Learning \and Object Detection \and Deep Learning    }

\section{Introduction}\label{sec:intro}
Object Detection is a critical computer vision problem involving both the localization and classification of objects within an image.
However, a fundamental limitation of Deep Learning systems is their inability to learn incrementally over time, as they are prone to Catastrophic Forgetting \cite{mccloskey1989catastrophic, goodfellow2013empirical} when exposed to new data.
Continual Learning (CL) is a research field focused on enabling models to learn new information while retaining previously acquired knowledge.
Applications such as robotics \cite{pasti2024tiny} and autonomous driving \cite{verwimp2023clad} can benefit greatly from models capable of adapting and learning continuously from streaming data.

\begin{figure*}[!t]
   \centering
   \includegraphics[width=\linewidth, trim = 0 0 0 0]{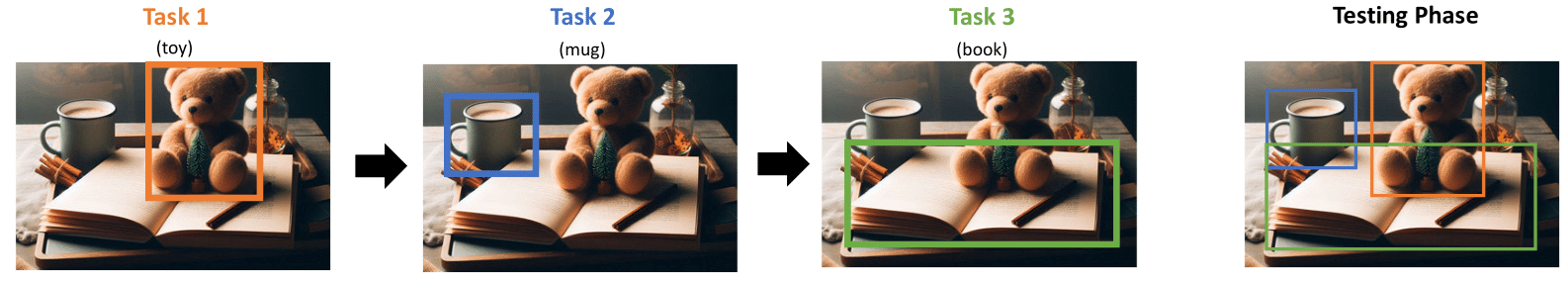}
   \caption{Continual Learning for Object Detection pipeline. The model learns to detect new classes at each incremental training stage (task). However, the scenario is challenging due to the missing annotations problem. The example in the current task might have been learned as background in previous tasks (e.g. the class toy is shown in Task 1 but not in Task 2 or 3.).
   }
   \label{fig:CLOD_scheme}
\end{figure*}

Most CL studies for image classification, have shown that replay-based approaches are the most effective method to achieve strong results \cite{yang2021benchmark,bagus2021investigation,tremonti2024empirical,pellegrini2020latent}. 
However, in the Continual Learning for Object Detection (CLOD) setting, replay-based approaches suffer from the \textit{missing annotations} problem \cite{menezes2023continual}.
For example, the image in Fig. \ref{fig:CLOD_scheme} contains the classes toy, mug, and book, but not all of these objects are consistently labeled across tasks.
This causes \textit{task interference} and \textit{memory efficiency} issues, making replay-based approaches underperform.

To address the limitations of replay approaches for CLOD, we propose a novel technique called Replay Consolidation with Label Propagation for Object Detection~(RCLPOD).
RCLPOD achieves superior performance on well-known benchmarks like COCO and VOC, demonstrating that replay-based approaches can perform as well or even better than distillation-based methods. 
RCLPOD improves the replay memory promoting class balance while also improving the quality of the ground truth of the stored samples through a technique called Label Propagation (see Fig. \ref{fig:OurApproach} for an overview of the RCLPOD framework).

Another limitation in the current CLOD landscape is that most studies are based on models like Faster R-CNN \cite{peng2020faster, liu2023augmented} or YOLO architectures such as YOLOv3 \cite{shieh2020continual} and v5 \cite{nenakhov2021continuous}.
However, these models differ significantly in terms of architecture design from recent YOLO versions.
This paper provides a benchmark for modern YOLO architectures by testing several CL methods using the anchor-free YOLOv8 architecture \cite{Jocher_Ultralytics_YOLO_2023}.
Therefore, this work can serve as a valuable benchmark for future research on modern YOLO architectures within the CLOD framework.

Overall, we make the following contributions:
\setlist{nolistsep}
\begin{itemize}[noitemsep] 
    \item We propose a novel approach called RCLPOD, which addresses key limitations of replay techniques in the CLOD setting.
    \item We conduct extensive experiments with RCLPOD, demonstrating its superior performance on the established VOC and COCO CLOD benchmarks.
    \item We benchmark our approach and other CL techniques using more modern versions of YOLO than those commonly evaluated in the literature.
\end{itemize}

The remainder of the paper is structured as follows. 
Sec.~\ref{sec:related_work} provides an overview of the CL literature focusing on CLOD.
In Sec.~\ref{sec:our_approach}, we describe our novel approach, RCLPOD.
Sec.~\ref{sec:experimental_setting} describes the experimental setting, including the considered metrics, evaluated scenarios, and experimental details.
In Sec.~\ref{sec:results}, we show and discuss the results.
Finally, Sec.~\ref{sec:conclusions_future_work} concludes this work by discussing limitations and future research directions.

\section{Related Work}
\label{sec:related_work}

\noindent Continual Learning aims to enable DL models to continuously accumulate knowledge over time without the need to re-train from scratch and with limited access to past data \cite{de2021continual}.
In particular, this framework has emerged to tackle the Catastrophic Forgetting problem \cite{mccloskey1989catastrophic, goodfellow2013empirical}; after a DL model is trained on one task and subsequently trained on another, it tends to forget the previously learned task.

\noindent Most of the literature on Continual Learning for Object Detection focuses on the Class Incremental Learning~(CIL) scenario, where each new task introduces new classes~\cite{menezes2023continual}. For example, Fig.~\ref{fig:CLOD_scheme} depicts a CLOD setting composed of three tasks where a new class must be learned each time. At the end of the training, the model should be able to provide predictions on all these classes.
Most of the works in CLOD literature were conducted on two-stage object detectors, whose time complexity scales linearly with the number of predicted objects in the scene. On the contrary, one-stage object detectors are known to be able to perform real-time object detection. In particular, recent papers focus on anchor-free architectures, with the most known probably being the modern versions of YOLO \cite{redmon2016you, Jocher_Ultralytics_YOLO_2023, wang2024yolov10}.

The first CL approach proposed for a CLOD scenario was proposed in~\cite{shmelkov2017incremental}. The authors propose to use the Learning without Forgetting~(LwF), a regularization-based approach, in the context of a two-stage object detector, Fast-R-CNN \cite{girshick2015fast}. 
LwF~\cite{li2017learning}, following the idea of knowledge distillation~\cite{hinton2015distilling}, forces the outputs of the training model, called student, to be similar to the corresponding outputs of a frozen version of the model trained on old tasks, called teacher.
Following this work, many subsequent in the literature are distillation-based approaches \cite{menezes2023continual}.
\\
\noindent However, all of them consider either two-stage object detectors \cite{girshick2015fast, ren2015faster} or one-stage object detectors like YOLOv3 \cite{redmon2018yolov3}, which are not anchor-free like fully convolutional one-stage object detector (FCOS) \cite{tian2020fcos} or more recent versions of YOLO like v8 \cite{ Jocher_Ultralytics_YOLO_2023}.

More recent works applied to anchor-free fully convolutional detectors~\cite{peng2021sid} revisit regularization-based algorithms for the FCOS object detector.
In particular, they address the noise introduced by the separate branch for regression outputs.
Building on this work, a solution for CLOD on the Edge called Latent Distillation was proposed \cite{pasti2024latent}.
However, these solutions are tailored FCOS-based architectures, which makes its adoption in modern YOLO architectures challenging.
\\
Another popular regularization-based approach is Pseudo Label \cite{guan2018learn}. Before training on the new task, the new task samples are passed through the old model, and its predicted labels are concatenated to the ground truth. 
However, like all distillation-based approaches, the main issue with pseudo-label is that performance remains high until images containing classes of the previous tasks are present in the new images. Otherwise, performance decreases quickly.
\\
\noindent The CLOD literature includes also Replay \cite{menezes2023continual}.
This method stores a portion of the old data in a memory buffer. While training on a new task, the new data is combined with the data from the memory buffer to maintain knowledge of old tasks \cite{hayes2021replay}.
According to the CL literature for image classification, Replay is one the most effective solutions for catastrophic forgetting \cite{yang2021benchmark, buzzega2021rethinking}.
\\
However, this is not the case in object detection, where \textit{task interference} arises caused by the \textit{missing annotations}. 
\\
\noindent To address these issues, the authors of~\cite{shieh2020continual} building on the YOLOv3 architecture proposed to mask the loss associated with new classes during the training of old samples and vice-versa for the new data.
However, the masking loss only partially solves the problem caused by the missing annotations.
This approach fails to fully leverage the potential information contained within the replay memory (\textit{memory efficiency} issue).

\noindent Contrary to the previously cited works, anchor-free one-stage object detectors are the most used in practical applications nowadays. Therefore, we propose to benchmark our method using the widely adopted YOLOv8, showing how our approach mitigates the problems that affect either distillation-based solutions or replay ones.

\section{Replay Consolidation with Label Propagation for Object Detection}
\label{sec:our_approach}
In this work, a novel approach is introduced called Replay Consolidation with Label Propagation for Object Detection (RCLPOD), aimed at improving Continual Learning in Object Detection by addressing task interference and enhancing memory replay.

\begin{figure*}[!t]
  \centering
  \begin{subfigure}[b]{0.9\linewidth}
    \centering
    \fbox{
    \includegraphics[width=\linewidth]{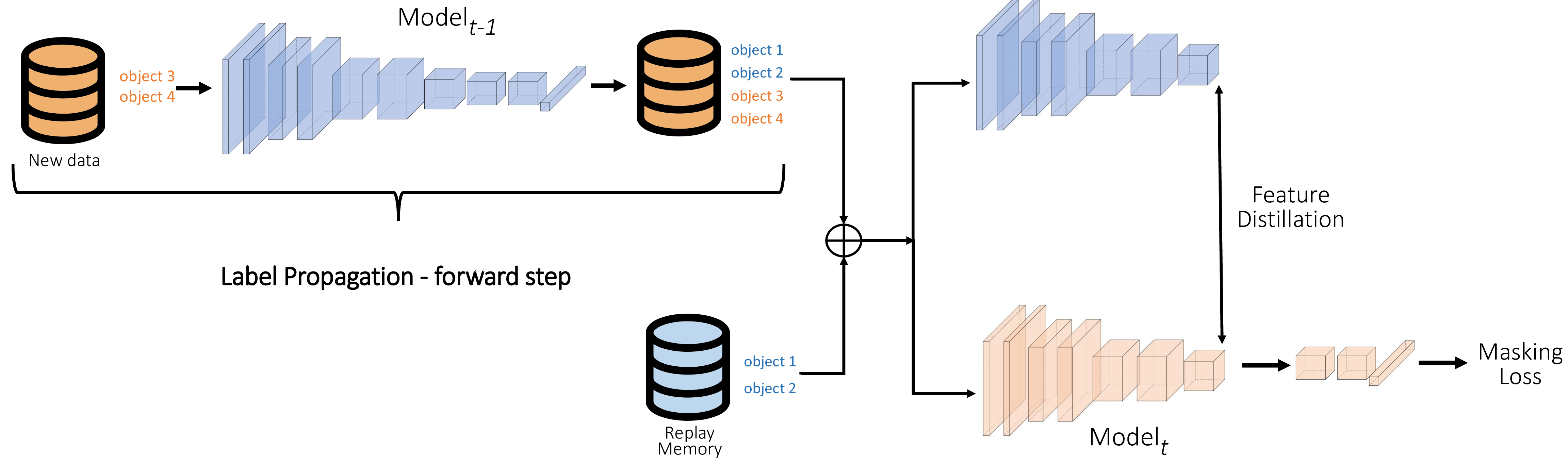}
    }
    \caption{RCLPOD approach during the training of a new task.}
    \label{fig:training_framework}
  \end{subfigure}
  \hfill
  \begin{subfigure}[b]{0.9\linewidth}
    \centering
    \fbox{
    \includegraphics[width=\linewidth]{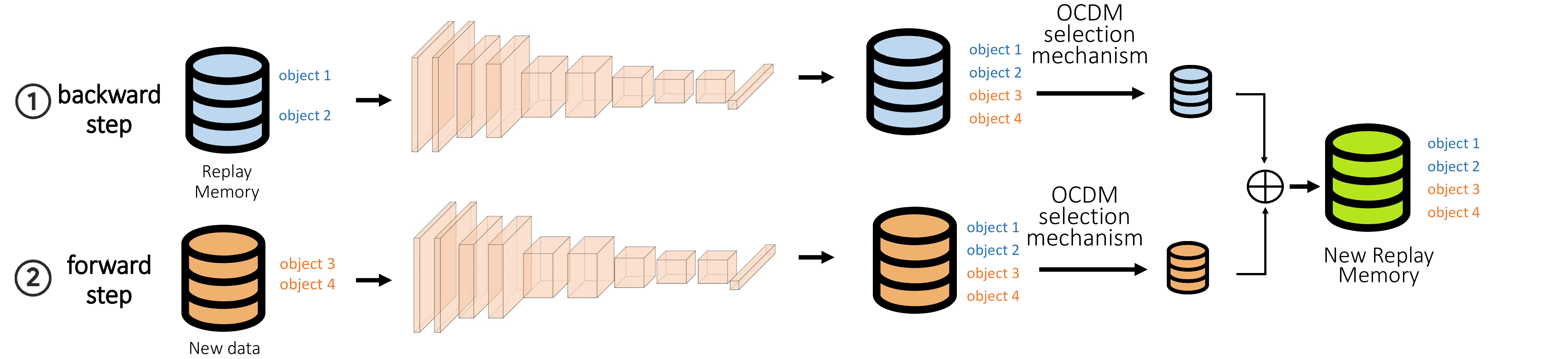}
    }
    \caption{Consolidation of the information inside the replay memory using the Label Propagation technique.}
    \label{fig:RCLP}
  \end{subfigure}
  \caption{Scheme of the RCLPOD method:
(a) During the training of the new task, the new samples  are first processed through the old model to generate pseudo-labels. Then they are combined with the samples from the replay memory to update the model, considering also additional losses like the Masking Loss and the one associated with feature distillation to reduce the drift of old intermediate representations. 
(b) Post-training procedure to update the replay memory. This procedure enhances the replay memory via label propagation. The backward step updates the old samples  with new knowledge, while the forward step integrates old knowledge into new task samples. Some of the new enriched samples are stored in the memory buffer.
}
  \label{fig:OurApproach}
\end{figure*}

\noindent One of the main issues of the CLOD setting is the \textit{missing annotations problem}.
In the CLOD setting, images seen in previous tasks may contain unlabeled classes that the model must learn as new classes in future tasks, causing missing annotations \cite{menezes2023continual}.
For example, the image in Fig.~\ref{fig:CLOD_scheme} has the classes person, dog, and car. However, this image has only one labeled class for each task. Then, this image could be saved multiple times in the replay memory, each time with different annotations based on the task.

\noindent This problem impacts negatively on the performance of Replay because it causes the \textit{task interference} and the \textit{memory efficiency} issues.
The \textit{task interference} arises between the current task samples and the replayed samples thar contain unlabeled instances of new classes.
This means that for those buffer samples, the ground truth interferes with the new class knowledge that the model is trying to learn

Moreover, even if such a problem was solved (e.g. through a Masking Loss \cite{shieh2020continual}), 
such an approach would still have the \textit{memory efficiency} issue.
This is because the full potential of the replay buffer is not realized, as each sample is stored in the memory buffer with ground truth labels limited to the specific classes of its original task (see Fig. \ref{fig:replay_memory}).

\subsection{Replay Memory with Label Propagation}
\label{subsec:label_propagation}
\begin{figure}[!h]
  \centering
  \begin{subfigure}[b]{0.3\linewidth}
    \centering
    \includegraphics[width=\linewidth]{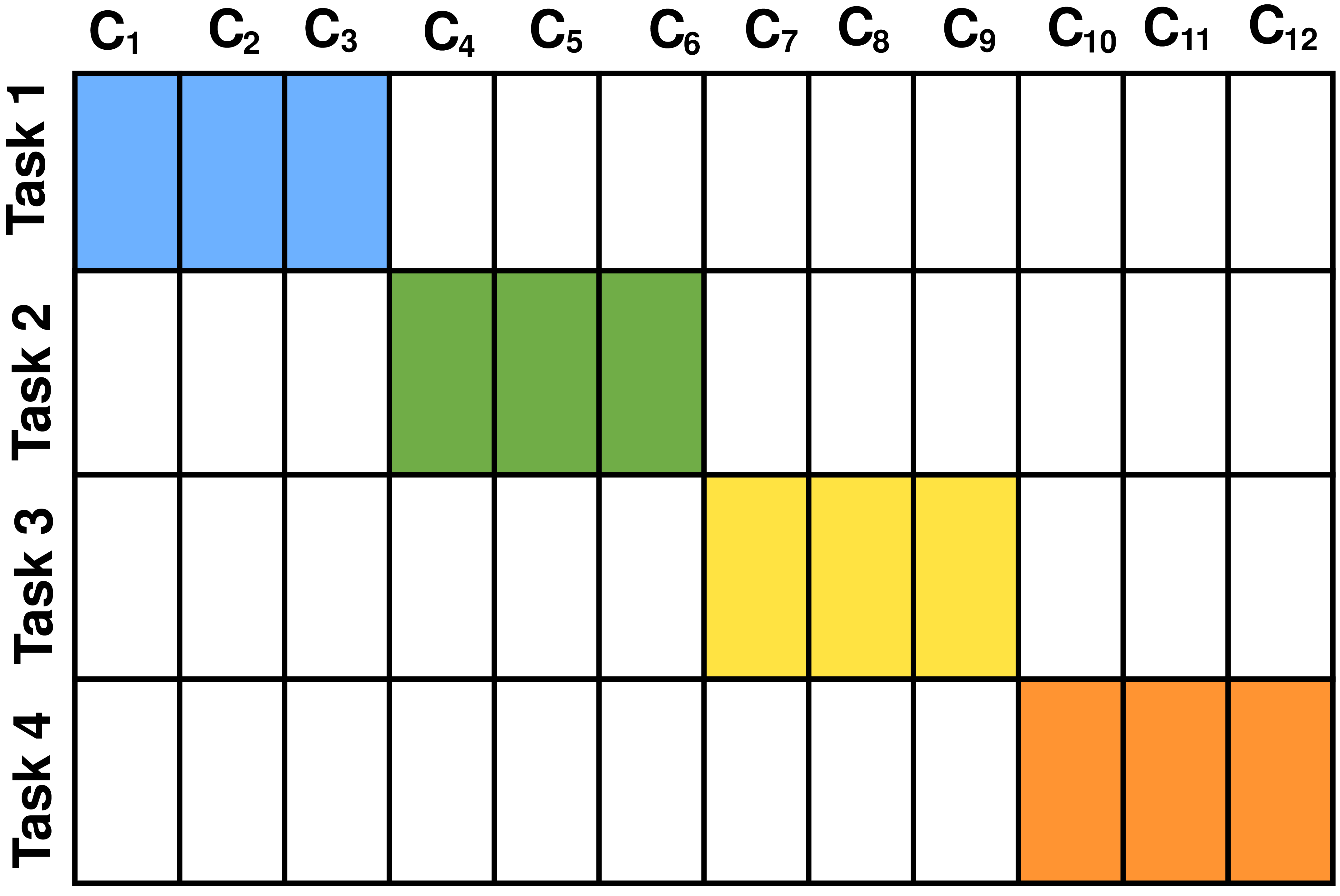}
    \caption{Replay Memory}
    \label{fig:replay_memory}
  \end{subfigure}
    \hspace{0.05\linewidth} 
  \begin{subfigure}[b]{0.3\linewidth}
    \centering
    \includegraphics[width=\linewidth]{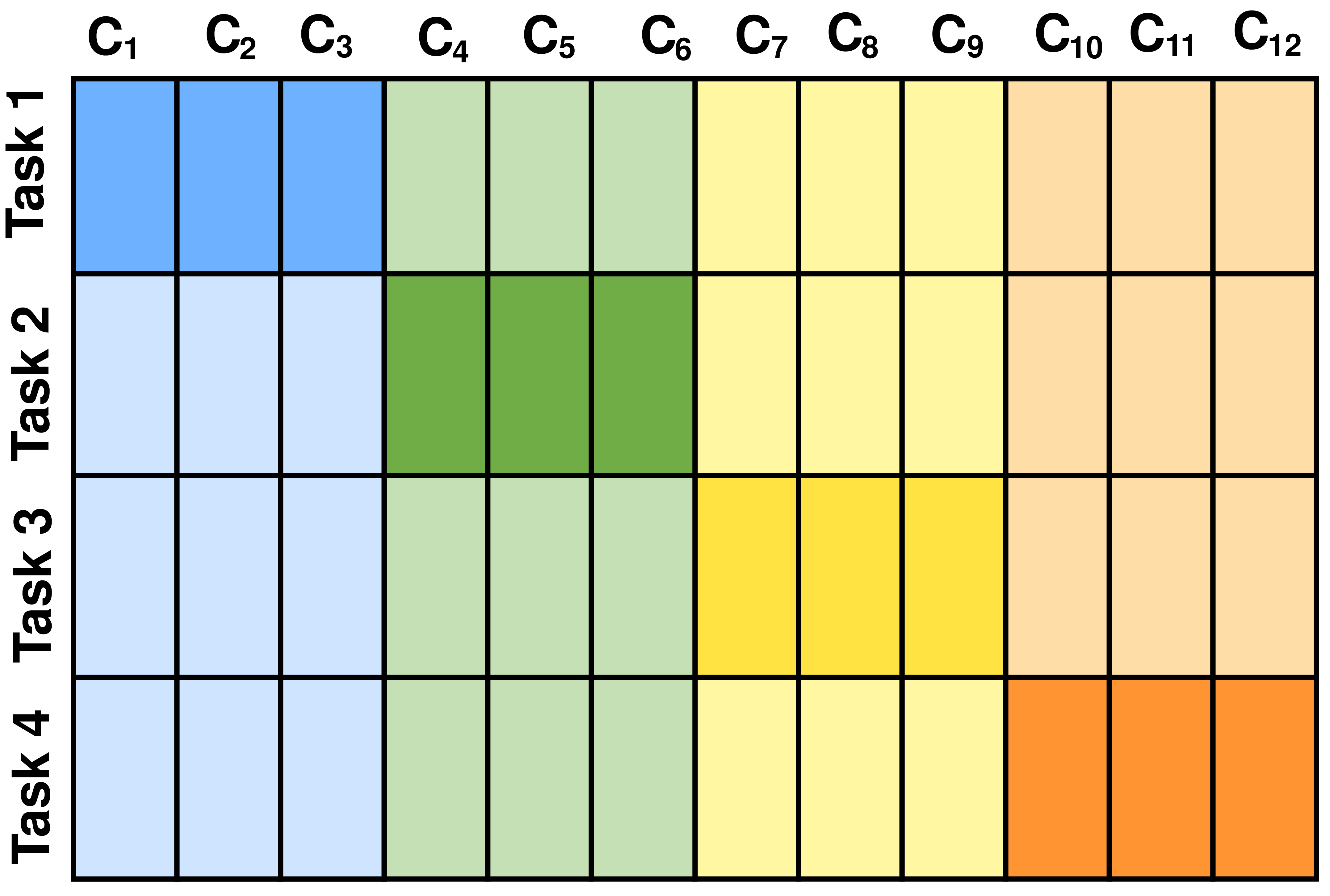}
    \caption{RCLPOD Memory }
    \label{fig:RCLP_memory}
  \end{subfigure}
  \caption{Comparison of the memory storage for Replay and RCLPOD. The vertical axis indicates the stored samples of each task, while the horizontal axis represents the objects associated with each class $c_i$. (a) Each task of the Replay memory has information only on the classes seen during its iteration. (b) Using the Label Propagation mechanism, the saved samples are more informative, containing knowledge of new and old classes.     
  }
  \label{fig:ComparisonMemory}
\end{figure}

To solve the discussed issues of the replay-based approaches in the CLOD setting,
we propose to implement a technique never considered before for CLOD called \textbf{Label Propagation (LP) mechanism}.
Originally proposed for the multi-label image classification setting, LP aims to enrich the replay memory by adding pseudo-labels to old and new samples \cite{ceccon2024multi} (see Fig. \ref{fig:ComparisonMemory}).
Adding pseudo-labels of previously learned classes to new task samples enriches the replay memory, allowing the model to retain knowledge of old classes while learning new ones.

\noindent While multi-label image classification and object detection differ significantly in terms of model architecture and training process, we argue that LP can be exploited to enhance the performance of CLOD.

We implement LP for CLOD considering the procedure structure into two different steps (see Fig. \ref{fig:RCLP}).
In the \textit{forward step}, information from old labels is added to the new data during training and subsequently into the replay buffer through a pseudo-labeling technique. 
In detail, let $D_t = \{X_t, Y_t\}$ be the data of the current task $t$ with associated a set of classes $C_t$ that appears in task $t$.
During the training of task $t$, for a sample $(x,y) \in D_t$, the forward step modifies the ground truth $y$ adding pseudo-labels of the classes $C_1, \ldots, C_{t-1}$ by leveraging the knowledge of the previously trained model $f_{\theta_{t-1}}$. Indeed, the model $f_{\theta_{t-1}}$ was trained with classes $C_1, \ldots, C_{t-1}$.
This method ensures that, after training on task $t$, the memory buffer retains samples with information relevant to all tasks up to and including task $t$.

On the other hand, the \textit{backward step} focuses on consolidating the new knowledge gained from training the model on the latest task into the old samples stored in the replay buffer. Specifically, using the new model trained on task $t$, pseudo-labels corresponding to the current task are added to the ground truth of old samples in the memory buffer.

\subsection{Balanced Replay Memory}
\label{subsec:masking_loss}

To improve the performance of RCLPOD, we propose to deviate from the classic selection mechanism adopted by Replay.
However, such a decision is sub-optimal since object detection datasets are normally unbalanced, like the case for VOC and COCO datasets.
Therefore, we propose enhancing the quality of the samples in the replay memory by ensuring a more balanced representation of each class.

\begin{figure}[!ht]
  \centering
  \begin{subfigure}[b]{0.35\linewidth}
    \centering
    \includegraphics[width=\linewidth]{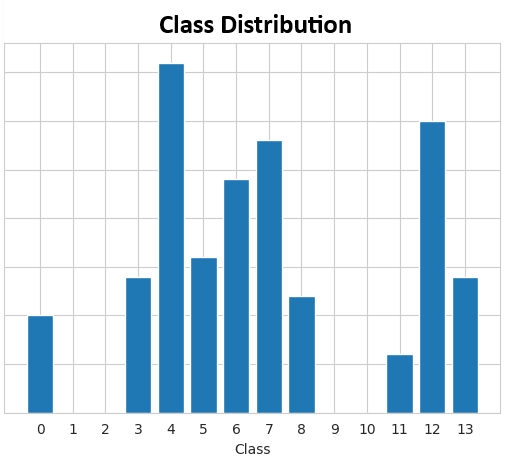}
    \caption{Replay Memory}
    \label{fig:label_freq_replay_memory}
  \end{subfigure}
     \hspace{0.05\linewidth} 
  \begin{subfigure}[b]{0.35\linewidth}
    \centering
    \includegraphics[width=\linewidth]{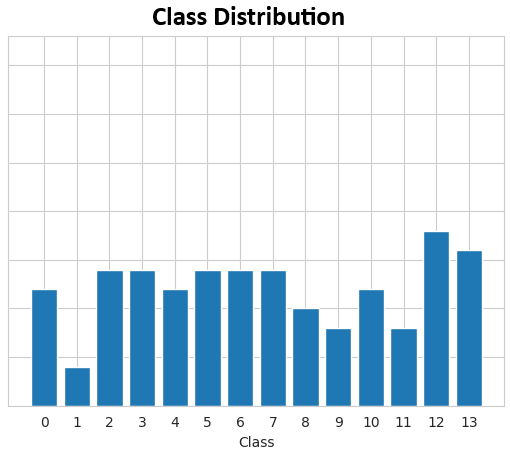}
    \caption{RCLPOD Memory }
    \label{fig:label_freq_RCLP_memory}
  \end{subfigure}
  \caption{An example comparing the class distributions in the memory buffer of Replay and RCLPOD. The vertical axis indicates the label frequency, while the horizontal axis represents the different labels. (a) On the left, Replay shows an unbalanced distribution. (b) At right, the distribution of RCLPOD is more balanced because of the selection mechanism.
  \label{fig:label_distribution_comparison}
  }
  \label{fig:label_freq_ComparisonMemory}
\end{figure}

\noindent Specifically, we propose to use a selection mechanism called OCDM \cite{liang2021optimizing}, which aims to have a memory buffer with a more balanced class distribution (see Fig. \ref{fig:label_freq_RCLP_memory}). 
This selection mechanism is advantageous because it is tailored for settings like CLOD, where each image is associated with multiple objects.
Giving higher priority to images with underrepresented classes helps achieve more balance and promotes class fairness.
Specifically, a greedy approach is applied which removes samples one by one based on how much they contribute to the target uniform distribution (see Sec. \ref{subsec:ocdm_pseudocode} for the pseudocode). 

\noindent In the CLOD context, OCDM and Label Propagation work in synergy to optimize the replay memory, as demonstrated by the conducted ablation study of Sec. \ref{subsec:ablation_study}.

If OCDM is applied without Label Propagation, it may need to select the same samples multiple times to achieve balanced class representation, which could limit the diversity of the samples in the buffer. However, with Label Propagation enhancing each sample with pseudo-labels for additional classes, the samples in memory become more informative. This allows OCDM to select a more varied set of samples, improving the overall representativeness of the replay memory while reducing redundancy.

\subsection{Masking overlapping objects} 
\label{subsec:masking_loss}
\begin{figure}[!h]
  \centering
   \includegraphics[width=0.35\linewidth, trim = 0 0 0 0]{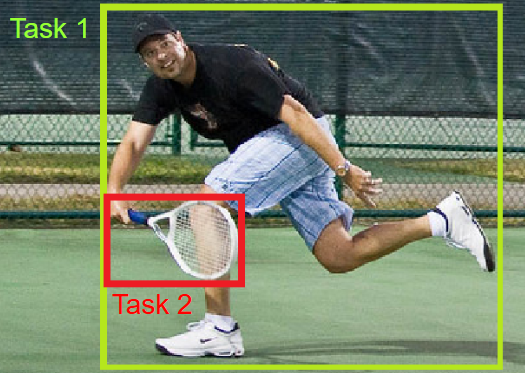}
   \caption{Example of task interference for YOLO architecture due to \textit{overlapping objects}.     
   Image from replay memory: class "tennis racket" is a new class, while class "person" is an old one. Since the only ground truth available is the one for the "person" class, any model prediction for the tennis racket would be penalized in the classification loss computation.}
   \label{fig:tennis_person}
\end{figure}

\noindent Label Propagation considerably reduces task interference in replay-based techniques.
However, the issue of handling new classes for replay memory samples when training on new tasks remains. On the other hand, since YOLO uses Task Alignment Learning (TAL) \cite{feng2021tood} for label-prediction assignment, the extent to which interference affects the performance depends on the saved images. In fact, YOLO’s assigner links each ground truth with at least one model prediction based on the intersection area. Thus, if a new-class object is distant from old-class objects, no ground truth is assigned to its prediction, and it doesn’t affect the loss. Consequently, when new-class objects rarely
overlap with old ones, interference is minimized. When instead the dataset presents frequent overlaps between objects, as in Fig. \ref{fig:tennis_person}, interference arises. We define this as the \textit{overlapping objects} issue.

\noindent To tackle this problem independently of the dataset, inspired by \cite{shieh2020continual},  we propose a custom masking loss approach for YOLOv8. In particular, we propose to ignore the contribution of new classes for the classification loss computation for every prediction associated to any sample saved in the memory buffer; recalling that, for a prediction assigned to a ground truth, the classification loss is given by $\mathcal{L}_{\text{cls}}=\sum_{i}\,\mathcal{L}_{\text{cls}}^i$, where $\mathcal{L}_{\text{cls}}^i$ is the usual Binary Cross Entropy for class $i$, the classification loss for any sample in the Replay memory is given by:

\begin{equation}
    \mathcal{L}_{\text{cls-mask}}=\sum_{i\not\in C_t}\,\mathcal{L}_{\text{cls}}^i
\end{equation}

\noindent where $C_t$ is the set of the new classes at task $t$. In the case of samples for the current task, the classification loss is the original one instead.

\subsection{Improving Stability}
\label{subsec:feature_distillation}
By employing the RCLPOD method as described above, we are able to provide more information to the replayed samples improving both stability and plasticity when learning a new CL task.
To further enhance model stability, we aim to minimize the drift of intermediate representations of previously learned objects. However, to balance the stability-plasticity trade-off, we consider low-level features, as they are more generalizable and can be shared across tasks.
Therefore, to improve the stability of the model and reduce forgetting, we employ a \textbf{Feature Distillation} (FD) technique \cite{li2024continual} by aligning the low-level features of the new and old models.

\noindent Specifically, for a YOLOv8 architecture, we perform distillation on the model's intermediate representations as follows:
Let our model $f(x) = g_\phi(h_\omega(x))$, where for an input $x$, $h$ gives the intermediate features, and $g$ produces the model output based on such features.
Formally, given an input sample $x$, the distillation loss on intermediate features is the following:
\begin{equation}\label{eq:feat_dist}
    \mathcal{L}_{\text{feat\_dist}} = || h_{\omega_t}(x) - h_{\omega_{t-1}}(x) ||_2^2
\end{equation}

\noindent where $\omega_t$ denotes the current model parameters while $\omega_{t-1}$ the ones belonging to the old model trained on the previous task. \\

However, the YOLOv8 architecture includes a neck in addition to the backbone $h_{\omega}$. Therefore, we compute the distillation loss on the intermediate features of the backbone and the neck. Recalling that the YOLOv8 backbone outputs three feature maps at multiple levels $C_3,C_4,C_5$ and the neck does the same $P_3,P_4,P_5$ \cite{lin2017feature, liu2018path}, the overall loss can be computed as:
\begin{equation}
    \mathcal{L} = \mathcal{L}_{\text{YOLO}}+ \lambda\mathcal{L}_{\text{feat\_dist}}
\end{equation}
\noindent where $\mathcal{L}_{\text{YOLO}}$ is the YOLOv8 loss, while $\mathcal{L}_{\text{feat\_dist}}$ is the feature distillation loss. Specifically, for YOLOv8, we consider the distillation of 6 feature maps $C_3, C_4, C_5, P_3, P_4, P_5$, 3 of the necks and 3 of the backbone, rewriting the generic equation \ref{eq:feat_dist} for YOLO as follows:
\begin{equation}\label{eq:feat_dist_yolo}
    \mathcal{L}_{\text{feat\_dist}} = \sum_{i=3}^6 || C_i^{\omega_{t}} - C_i^{\omega_{t-1}} ||_2^2
    + || P_i^{\omega_{t}} - P_i^{\omega_{t-1}} ||_2^2
\end{equation}

\section{Experimental Setting}\label{sec:experimental_setting}
\medskip
\subsection{Scenarios and Metrics}
\noindent Following previous works~\cite{shmelkov2017incremental, peng2021sid, menezes2023continual}, we test our proposed CLOD method on the 2017 version of the PASCAL VOC~\cite{everingham2010pascal} detection benchmark, which includes 20 different object classes, and on the Microsoft COCO challenge dataset~\cite{lin2014microsoft} which has 80 object classes. 
Following prior works \cite{shmelkov2017incremental, peng2021sid} we evaluate our solution on the most well-known CIL scenarios. 
Using the notation $N\text{p}M$, the first task ($i = 1$) consists of the first $N$ classes in the list (e.g., in alphabetical order), while each subsequent task ($i > 1$) consists of the classes from $N + (i-2) \cdot M$ to $N + (i-1) \cdot M - 1$.
For example, given the VOC dataset with 20 classes, the CL scenario $15\text{p}1$~(read 15 plus 1) consists of 6 tasks: the first one with the first 15 object classes, the second one with the 16-th class, the third one with the 17-th class and so on. 
Another example is 15p5 with the first task being trained with 15 classes and the second one with 5 classes.

\noindent The CL scenarios for VOC are $15\text{p}1$, $15\text{p}5$, $10\text{p}10$ and $19\text{p}1$, while for COCO $40\text{p}40$ and $40\text{p}10$. 
The longest streams of tasks are VOC $15\text{p}1$ and COCO $40\text{p}10$ while the others are relatively short.

\noindent As in~\cite{shmelkov2017incremental} to evaluate the performance of YOLOv8, we report the mean average precision (mAP) at the end of the training.
Specifically, considering a 0.5 IoU threshold (mAP$^{50}$) for VOC and the mAP weighted across different IoU from 0.5 to 0.95 (mAP$^{50-95}$) for COCO. 

\subsection{YOLO training details}

\noindent To evaluate how the CL performances are affected by the model size, in our study we consider two different versions of YOLOv8: YOLOv8n with 3.2M parameters and YOLOv8m with 25.9M parameters.
We initialize the backbone parameters with the ones pre-trained on Image-Net, available on \cite{Jocher_Ultralytics_YOLO_2023}.
\noindent In Table \ref{tab:hyper} the hyper-parameters used for training are reported. In particular, by following \cite{Jocher_Ultralytics_YOLO_2023, wang2024yolov10}, we use SGD with Nestorov momentum and weight decay. For both the learning rate and the momentum we have 3 warmup epochs, while just for the learning rate we employ a linear decay scheduling from $10^{-2}$ to $10^{-4}$. In all the experiments we set the number of epochs per task to 100, which is a suitable value for reaching convergence.

\subsection{CL strategies details}

\noindent We compare the following methods against RCLPOD:
\textit{Joint training, Fine-tuning, Replay, OCDM, LwF, and Pseudo-Label}.
In particular, while Replay, LwF, and Pseudo-label were already evaluated in the CLOD setting, this is the first time, to the best of our knowledge, that the OCDM approach is implemented and tested in the CLOD setting.
Join Training is considered as an upper bound and corresponds to the model performances when trained on the entire dataset with all the classes. 
Fine-tuning is used as a lower bound; each task is trained sequentially, and no CL technique is employed to avoid catastrophic forgetting.\\
\noindent Regarding the Replay methods, the memory has a small capacity fixed to around 5\% of the entire dataset, namely 800 images for PASCAL VOC and 6,000 for COCO. For OCDM, we use the same memory size as all the other replay-based approaches. Moreover, we provide additional results for the replay-based approaches in Sec. \ref{subsec:mem} by changing the memory size to evaluate its impact on performance.

\noindent For LwF, by following \cite{shmelkov2017incremental}, we set the distillation loss weight $\lambda$ to 1.
For Pseudo-Labeling, to ensure consistency with inference, we set the classification threshold to $0.5$ and the IoU threshold, for Non-maximum Suppression, to $0.7$.
For RCLPOD we use the same hyper-parameters of Pseudo-Labels and Replay. As for LwF, we set the gain for the feature distillation loss to $\lambda=1$.

\begin{table*}[th]
\centering
\begin{tabular}{l|cccc|cc}
\toprule
\multirow{2}{*}{\textbf{}} & \multicolumn{6}{c}{\textbf{Scenarios}} \\ 
\cmidrule{2-7} & \multicolumn{4}{c}{\textbf{VOC (mAP$_{50}$)}} & \multicolumn{2}{|c}{\textbf{COCO (mAP$_{50-95}$)}}       \\ \midrule
\textbf{CL Method}         &  \textbf{10p10} &  {\textbf{19p1}} &  {\textbf{15p5}} &  {\textbf{15p1}} &  {\textbf{40p40}} & \textbf{40p10} \\  
\textbf{Joint Training}    &  {78.5}          &  {78.5}         &  {78.5}              &  {78.5}         &  {37.3}          & 37.3          \\  
\textbf{Fine-Tuning}       &  {36.4}              &  {16.5}              &  {16.4}              &  {3.0}              &  {13.6}               &        2.9        \\ 
\textbf{Replay \cite{hayes2021replay}}            &  {54.5}               &  {\underline{60.8}}              &  {56.9}              &  {39.0}              &  {14.1}               &           8.3    \\    
\textbf{LwF \cite{li2017learning}}               &  {57.3}               &  {60.0}              &  {57.0}              &  {41.5}              &  {19.0}               &          \underline{16.2}     \\  
\textbf{OCDM \cite{liang2021optimizing}}              &  {56.5}               &  {59.9}              &  {54.7}              &  {\underline{49.6}}              &  {16.7}               &          11.0      \\  
\textbf{Pseudo-label \cite{guan2018learn}}      &  {\underline{71.4}}               &  {46.6}              &  {\underline{59.7} }            &  {16.3}              &  {\textbf{24.4}}               &        \underline{16.2}        \\  \midrule
\textbf{RCLPOD (ours)}     &  {\textbf{72.5}}      &  {\textbf{68.4}}     &  {\textbf{67.7}}              &  {\textbf{56.1}}     &  {\underline{24.1}}      &   \textbf{20.0}             \\  
\bottomrule
\end{tabular}
\caption{Results for YOLOv8n. Each column represents one of the studied scenarios based on VOC and COCO datasets. Each row represents a different tested CL technique. In bold the best method and underlined the second best method. As in CLOD literature, mAP50 is used for VOC-based scenarios and mAP 50-95 for COCO-based scenarios. }
\label{tab:results}
\end{table*}

\section{Results}
\label{sec:results}

\noindent In this section, we discuss the outcomes of each method on the VOC and COCO CLOD benchmarks. Table \ref{tab:results} reports the results after all tasks have been completed for YOLOv8n. Additional results for YOLOv8M are provided in Sec. \ref{subsec:YOLOv8M}.
In Sec. \ref{subsec:res2tasks}, we discuss the results obtained in the 2-tasks scenarios, namely 15p5, 10p10, 19p1, and 40p40. 
In Sec. \ref{subsec:res_n_tasks}, we consider the more challenging scenarios 15p1 and 40p10. In Sec. \ref{sec:stab-plas}, we compare the different methods in terms of stability-plasticity. 

\subsection{Two tasks scenarios}
\label{subsec:res2tasks}
\noindent The results for all the two task scenarios are reported in Tab.~\ref{tab:results}.
Specifically, scenarios 10p10, 19p1, and 15p5 for the VOC dataset and scenario 40p40 for the COCO dataset.

\noindent Replay shows good performance on the VOC dataset, particularly in the 10p10 and 19p1 scenarios, where it achieves mAP scores of 54.5 and 60.9, respectively. This indicates that Replay can effectively retain knowledge from earlier tasks when applied to simpler datasets like VOC.
When comparing the Replay method with OCDM, we can observe similar results except for the scenario COCO 40p40, which has a slight improvement.

\noindent Notably, OCDM performs better than Replay in the long 15p1 scenario with a mAP of 50.2, highlighting its effectiveness in handling challenging incremental tasks. 

\noindent Regarding Pseudo-label, we can observe its effectiveness for most of the scenarios except for the 19p1, where the gap to RCLPOD is significant. We suppose this is due to the low number of old-class objects in the images for the new tasks; the lower the number of classes for the second task, the less overlap between tasks we have (see the plots of Fig. \ref{fig:plot_results_voc_19p1} in the Appendix). On the contrary, this is not the case for the 40p40 scenario, where Pseudo-label, despite its simplicity, reaches the best mAP.
Even if the overall mAP is higher than the RCLPOD one, a higher forgetting is observed, while plasticity brings that higher final result. For more details, we refer to Sec. \ref{subsec:YOLOv8M} Appendix, where we compare RCPLOD and Pseudo-label, the two best methods in the 40p40 scenario, in the case of the bigger model YOLOv8m. 

\noindent For what concerns LwF, the results obtained show that this method can obtain good performance in many scenarios. However, we can observe the  model plasticity is significantly reduced. This is due to the noisy regression output of the teacher, then the loss between the teacher and the student output prevents the model from acquiring new knowledge. For a more detailed discussion, we refer to Sec. \ref{ap:lwf} in the Appendix.

\begin{figure*}[th]
  \centering
  \begin{subfigure}[b]{0.49\linewidth}
    \centering
    \includegraphics[width=\linewidth]{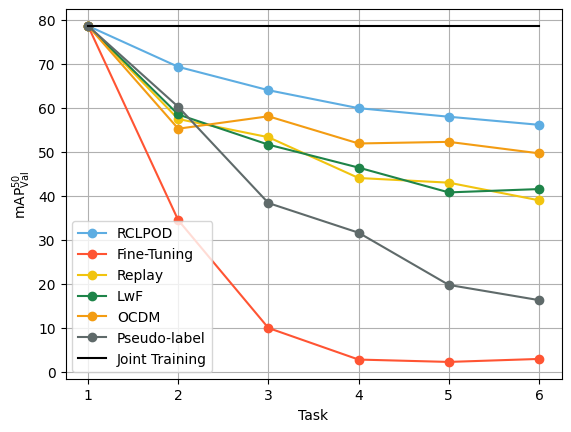}
    \caption{Scenario VOC 15p1 with the mAP50.}
    \label{fig:results_voc15p1}
  \end{subfigure}
    \hfill
  \begin{subfigure}[b]{0.49\linewidth}
    \centering
    \includegraphics[width=\linewidth]{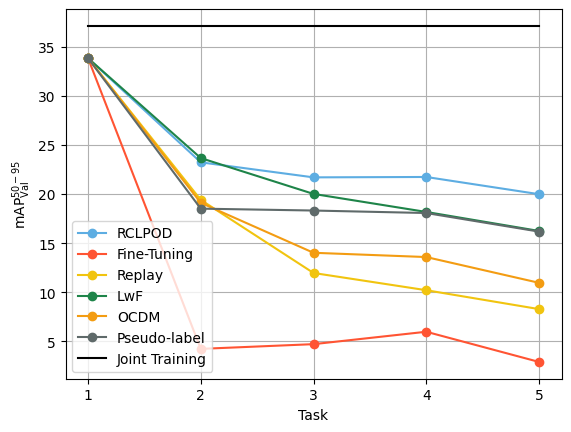}
    \caption{Scenario COCO 40p10 with mAP 50-95. }
    \label{fig:results_voco40p10}
  \end{subfigure}
  \caption{Results obtained for VOC 15p1 at left and COCO 40p10 at right. Ours method RCLPOD performs best in both scenarios. 
  }
  \label{fig:plot_results}
\end{figure*}

\subsection{Scenarios VOC 15p1 and COCO 40p10}
\label{subsec:res_n_tasks}
\noindent Here, we discuss the two longer and more challenging streams: VOC 15p1 and COCO 40p10.
COCO 40p10 given its size and variety in terms of images, makes it a more complex stream than the VOC dataset.
However, at the same time, the task overlapping (number of images in common among tasks) is very low in VOC $15\text{p}1$ compared to COCO 40p10 (see plots in Sec. \ref{subsec:label_distribution_scenarios}), making the analysis of this scenario very meaningful.
This is due to the high number of classes per task in COCO, while by adding a single class each time, VOC $15\text{p}1$ has a lower degree of task overlap.

\noindent As highlighted in Tab \ref{tab:results}, Pseudo-Label works very well in a complex scenario like COCO 40p10.
This is partially due to the very high intersection between tasks in terms of images, which allows the model to maintain knowledge of previous classes as it implicitly revisits them in the new task.
However, in cases where the intersection between tasks is very low, the Pseudo-Label performance is expected to be much worse.
This is demonstrated by the VOC 15p1 scenario, where performance is low as only one new class is shown, and the intersection with previous tasks is much lower compared to COCO 40p10, where ten new classes are added each time. 

\noindent Therefore, in practice, we have this interesting duality aspect where Replay for VOC 15p1 performs better than Pseudo-Label and vice versa for the COCO 40p10 scenario.
Unlike Pseudo-label, interference between tasks in Replay is a problem because it implies saving the same image in memory several times with different labels, which causes several issues, including an underexploitation of memory.
OCDM, despite having the same problem of missing annotations, is rewarded by carefully selecting the samples to save, which improves the final performance from 8 for Replay to 11 for OCDM in the COCO 40p10.

\noindent Then RCLPOD, which enhances the replay memory by improving the ground truth associated with the samples, receives a significant boost, achieving 20 in the COCO 40p10.
This proves how effective the correct management of the information about the samples stored in the replay memory can be.
\noindent Finally, even if LwF reaches good performance in both scenarios, the same problem highlighted for shorter tasks, namely, that the model achieves poor performance for new tasks, is observed.

\subsection{Stability-Plasticity Balance}\label{sec:stab-plas}

\noindent In Tab. \ref{tab:oldnew} we report a comparison in terms of stability and plasticity in the two more challenging scenarios. In particular, at each new task, we measure the mAP for old classes, namely classes that appear in the old task, and the mAP for new classes; at the end of the last task, we compute the average both for old classes and new classes, and we report the results in Tab. \ref{tab:oldnew}. By doing so, we measure the average performance of any method both in terms of forgetting (old) and plasticity (new). As highlighted by the results, our method is prone to reducing forgetting, and in the COCO 40p10 scenario, it is the best method for balancing both stability and plasticity. However, the VOC15p1 scenario shows there is still room for improvement in terms of plasticity.
Regarding LwF, we discuss additional results regarding the stability-plasticity trade-off in \ref{ap:lwf}.

\begin{table*}[!htb]
\centering
\begin{tabular}{l|cccc|cccc}
    \toprule
    \multirow{2}{*}{\bf CL Method} &&
      \multicolumn{3}{c}{{\bf VOC 15p1} ({\bf mAP}$_{50}$)}&\multicolumn{4}{c}{{\bf COCO 40p10} ({\bf mAP}$_{50-95}$)}\\
      && {old} & {new} &all& {old} & {new}&all\\
      \midrule
      \textbf{Fine-Tuning}&&9.0&38.3&3.0&0.0&{\bf28.7}&2.9\\
      \textbf{Replay}&&47.0&53.8&39.0&12.7&11.9&8.3\\
      \textbf{LwF}&&48.8&31.6&41.5&20.5&15.4&16.2\\
      \textbf{OCDM}&&53.4&{\bf 54.9}&49.6&14.1&16.5&11.0\\
    \textbf{Pseudo-label} && 32.9 & 41.3 &16.3&16.3&26.4&16.2\\
   \textbf{RCLPOD (Ours)} && {\bf62.6}&43.0&\textbf{56.1} & {\bf 21.6}&22.6&\textbf{20.0}\\
    \bottomrule
  \end{tabular}
\caption{Results for YOLOv8n for VOC 15p1 and COCO 40p10 scenarios. "old" is the average mAP for the old classes over all the tasks, "new" is the average mAP for the new classes over all the tasks.}
  \label{tab:oldnew}
\end{table*}

\section{Conclusions and Future Work}
\label{sec:conclusions_future_work}
\noindent 
To improve the performance of Continual Object Detection, we propose a novel approach called RCPLOD.
Our method enhances the replay memory through class distribution balancing and Label Propagation, using existing data more effectively while solving the missing annotations and the task interference problem.
RCLPOD outperforms existing techniques on well-established benchmarks such as VOC and COC.

\noindent The new enhanced memory strategy described in this work provides an alternative to distillation-based approaches and can be reapplied to other architectures.
In addition, our algorithm doesn't occupy more memory than other approaches like Replay and LwF.
Similarly, while the computing could be slightly higher than Replay due to the procedure to enhance the replay memory, it is still a lightweight approach since it just needs to modify the information stored in the replay memory.
\noindent Moreover, our approach is developed to work with modern architectures like YOLOv8, making it suitable for dynamic, real-world applications such as autonomous driving \cite{verwimp2023clad} and robotics \cite{pasti2024tiny}, where continuous learning and resource efficiency are essential.

As future research, we believe that our approach can be used as a base for the proposal of new advancements in replay-based approaches.
Indeed, the Label Propagation mechanism tested in the CLOD scenario for the first time proves its efficacy, making it a suitable base to build upon since it facilitates more stable and memory-efficient models.
In particular, the base of our strategy is designed to work independently of specific detector architectures. The technique manages memory and labeling in ways that are generally compatible with any object detection model using replay-based continual learning, simplifying the extension to other architectures.
In addition, future research should give more attention to longer and more complex streams like COCO40p10 and beyond to better represent the real performances of the CL techniques in the CLOD setting.
Following this direction streams closer to realistic scenarios should be considered to better represent the challenges of Continual Object Detection.

\bibliographystyle{unsrt}  
\bibliography{references}

\clearpage
\setcounter{page}{1}

\appendix
\section{Appendix}
\subsection{Further training hyper-parameters details}
Table \ref{tab:hyper} presents the hyper-parameters used for training YOLO. Regarding data augmentation, we follow \cite{Jocher_Ultralytics_YOLO_2023, wang2024yolov10}.

\begin{table}[h!]
  \begin{center}
    \caption{Hyper-parameters for YOLOv8}
    \label{tab:hyper}
    \begin{tabular}{cc} 
    \hline
      \textbf{Hyper-parameter} &  \\
      \hline
      epochs & $100$\\
      optimizer & SGD \\
      batch size & $8$\\
      momentum & $0.937$ \\
      weight decay& $5\times 10^{-4}$\\
      warm-up epochs&$3$\\
      warm-up momentum& $0.8$\\
      warm-up bias learning rate & $0.1$\\
      initial learning rate & $10^{-2}$\\
      final learning rate & $10^{-4}$\\
      learning rate scheduling& linear decay\\
      box loss gain & $7.5$\\
      classification loss gain & $0.5$\\
      DFL loss gain & $1.5$\\
      \hline\\
    \end{tabular}
  \end{center}
\end{table}

\subsection{Ablation study}
\label{subsec:ablation_study}

\begin{table}[]
\centering
\begin{tabular}{c|cccccc}
\toprule
                 & \textbf{Replay} & \textbf{OCDM} & \textbf{LP} & \textbf{FD} & \textbf{Mask} & \textbf{mAP} \\ \midrule
\textbf{Model 1} &\ding{51}              &\ding{55}          &\ding{55}            &\ding{55}          &\ding{55}            & 14.1           \\
\textbf{Model 2} &\ding{51}              &\ding{51}          &\ding{55}            &\ding{55}          &\ding{55}            & 16.7           \\ 
\textbf{Model 3} &\ding{51}              &\ding{51}          &\ding{51}            &\ding{55}          &\ding{55}            &   19.7       \\ 
\textbf{Model 4} &\ding{51}              &\ding{51}          &\ding{51}            &\ding{51}          &\ding{55}            &   20.1      \\ 
\textbf{Model 5} &\ding{51}              &\ding{51}          &\ding{51}            &\ding{51}          &\ding{51}            & 24.1           \\ 
\textbf{Model 6} &\ding{51}              &\ding{55}          &\ding{51}            &\ding{51}          &\ding{51}            & 22.7           \\  \bottomrule
\end{tabular}
\caption{Ablation study performed to study the effect of each component in RCLPOD, tested on COCO 40p40. LP is the Label Propagation mechanism, OCDM indicates the selection mechanism, FD represents the Feature Distillation component, and Mask represents the masking component.}
\label{tab:ablation_study}
\end{table}

\noindent Here, we discuss the contribution of each part of our RCLPOD approach to the final performance.
Specifically, we identify four main parts.
The Label Propagation (LP) mechanism described in Sec. \ref{subsec:label_propagation} to enhance the replay memory. 
The selection mechanism (OCDM) to balance the class distribution of the replay memory that promotes class fairness.
The masking (Mask) introduced in Sec. \ref{subsec:masking_loss} to reduce the task interference in the case of overlapping objects for the YOLO architecture. 
Finally, Feature Distillation (FD) is applied to low-level representations to improve the stability and reduce forgetting described in Sec. \ref{subsec:feature_distillation}.

\noindent Tab. \ref{tab:ablation_study} reports the ablation results.
 First, we compare Replay (Model 1) with OCDM (Model 2). Even in a short CL scenario like COCO 40p40, OCDM performs better than Replay, showing the positive effect of balancing the Replay memory. By adding the Label Propagation (LP) mechanism (Model 3), a significant improvement of 3 mAP is obtained. In fact, by doing so, we both reduce the interference problem, and we fully utilize the potential of the replay memory. 
Feature Distillation (Model 4) brings an additional improvement of 0.4 mAP.
Once we add the masking component (Model 5), using all the components of RCLPOD, we get a further increase of 4 mAP. In fact, as shown in Fig.\ref{fig:freq_40p40}, the important overlap of many classes between tasks brings interference. Therefore, when a Replay memory is employed, masking is needed even in the case of modern versions of YOLO.
Finally, we test OCDM once all the other components are used. By removing OCDM (Model 6), a 1.4 mAP drop is observed, showing the importance of balancing the replay memory in reducing forgetting, particularly for longer streams such as VOC 15p1 and COCO 40p10.

\subsection{Results with YOLOv8m}
\label{subsec:YOLOv8M}

\noindent As shown in Tab. \ref{tab:results}, Pseudo-label performs slightly better than RCLPOD. However, as shown in \ref{tab:yolov8m}, Pseudo-label performs worse than RCLPOD in terms of forgetting, while it outperforms the latter by plasticity. \\
As proved in \cite{ramasesh2021effect}, robustness to forgetting improves with scale of model size. Therefore, we also investigate the effect of the model size in the COCO 40p40 scenario. As presented in \ref{tab:yolov8m}, RCLPOD performs better than Pseudo-Label, and the gap in terms of plasticity is lower than in the case of YOLOv8n. This fact shows that once the model size increases, the intrinsic plasticity of the model allows one to acquire new knowledge independently by the CL strategy applied to prevent forgetting.

\subsection{Label Distribution for each VOC and COCO scenario}
\label{subsec:label_distribution_scenarios}

\begin{figure*}[ht!]
  \centering
  \begin{subfigure}[b]{0.49\linewidth}
    \centering
    \includegraphics[width=\linewidth]{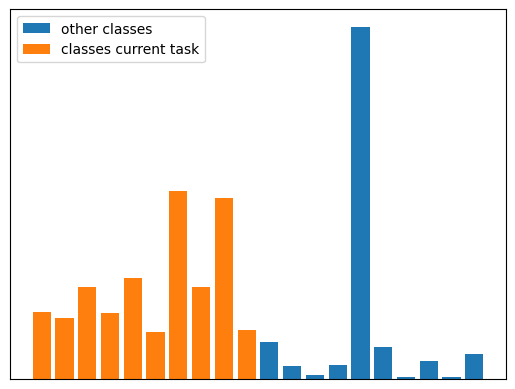}
    \caption{Task 1}
        \label{fig:stats10p10-1}
  \end{subfigure}
    \hfill
  \begin{subfigure}[b]{0.49\linewidth}
    \centering
    \includegraphics[width=\linewidth]{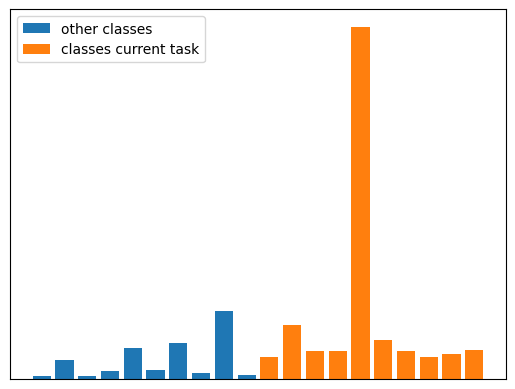}
    \caption{Task 2}
    \label{fig:stats10p10-0}
  \end{subfigure}
  \caption{Classes distribution over tasks for VOC 10p10. In blue the classes without labels.
  }
  \label{fig:freq:10p10}
\end{figure*}

\begin{figure*}[ht!]
  \centering
  \begin{subfigure}[b]{0.49\linewidth}
    \centering
    \includegraphics[width=\linewidth]{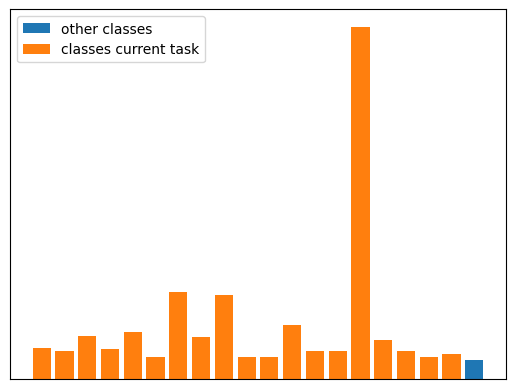}
    \caption{Task 1}
        \label{fig:stats10p10-1}
  \end{subfigure}
    \hfill
  \begin{subfigure}[b]{0.49\linewidth}
    \centering
    \includegraphics[width=\linewidth]{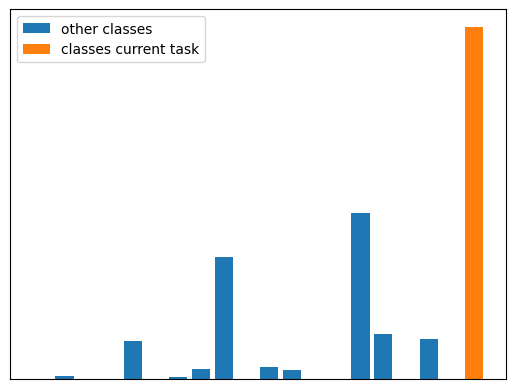}
    \caption{Task 2}
    \label{fig:stats10p10-0}
  \end{subfigure}
  \caption{Classes distribution over tasks for VOC 19p1. In blue the classes without labels.
  }
  \label{fig:plot_results_voc_19p1}
\end{figure*}

\begin{figure*}[ht!]
  \centering
  \begin{subfigure}[b]{0.49\linewidth}
    \centering
    \includegraphics[width=\linewidth]{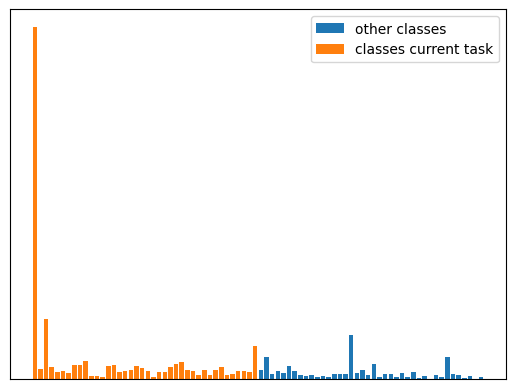}
    \caption{Task 1}
        \label{fig:stats10p10-1}
  \end{subfigure}
    \hfill
  \begin{subfigure}[b]{0.49\linewidth}
    \centering
    \includegraphics[width=\linewidth]{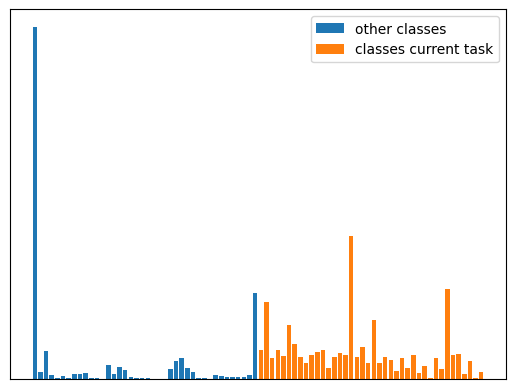}
    \caption{Task 2}
    \label{fig:stats10p10-0}
  \end{subfigure}
  \caption{Classes distribution over tasks for COCO 40p40. In blue the classes without labels.
  }
  \label{fig:freq_40p40}
\end{figure*}

\begin{figure*}[h]
\centering
\begin{tabular}{cccc}
\includegraphics[width=0.32\textwidth]{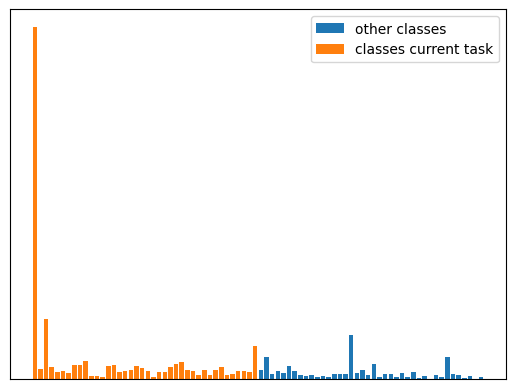} &
\includegraphics[width=0.32\textwidth]{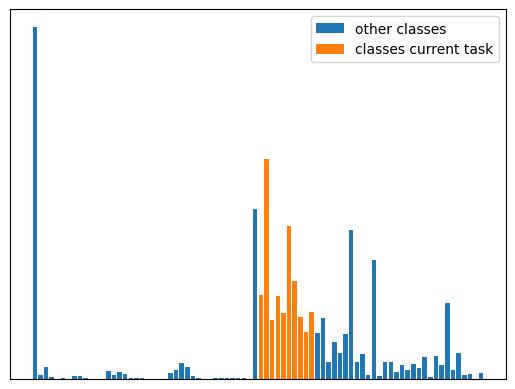} &
\includegraphics[width=0.32\textwidth]{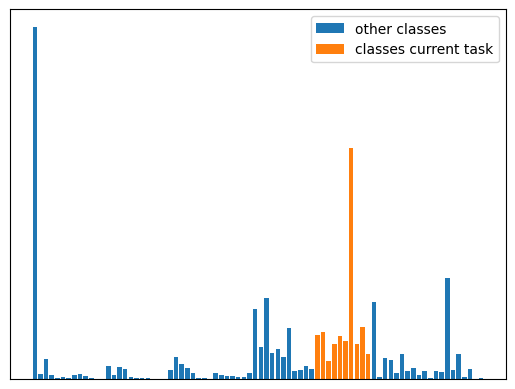} \\
Task 1  & Task 2 & Task 3\\[6pt]
\end{tabular}
\begin{tabular}{cccc}
\includegraphics[width=0.32\textwidth]{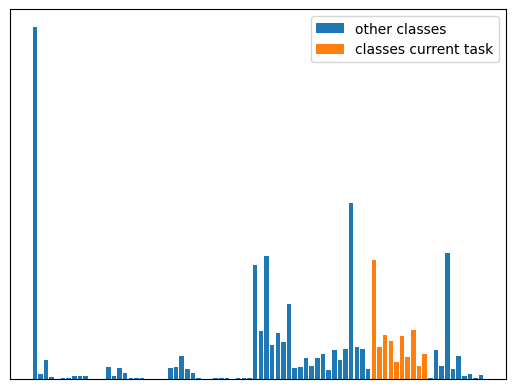} &
\includegraphics[width=0.32\textwidth]{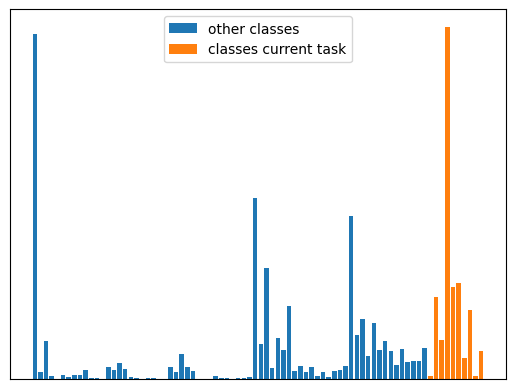} \\
Task 4  & Task 5  \\[6pt]
\end{tabular}
\caption{ Classes distribution over tasks for COCO 40p10. In blue the classes without labels.}
\label{fig:Name}
\end{figure*}

\begin{figure*}[htb]
    \centering 
\begin{minipage}[t]{.35\textwidth}
\begin{subfigure}{\textwidth}
  \includegraphics[width=\linewidth]{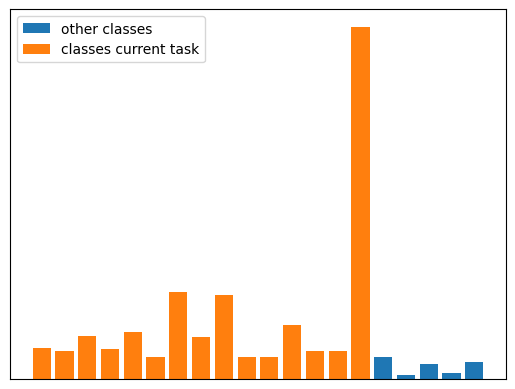}
  \caption{Task 1}
  \label{fig:1}
\end{subfigure}\hfil 
\begin{subfigure}{\textwidth}
  \includegraphics[width=\linewidth]{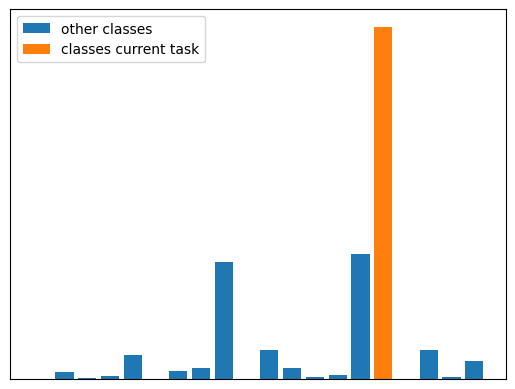}
  \caption{Task 2}
  \label{fig:2}
\end{subfigure}\hfil 
\begin{subfigure}{\textwidth}
  \includegraphics[width=\linewidth]{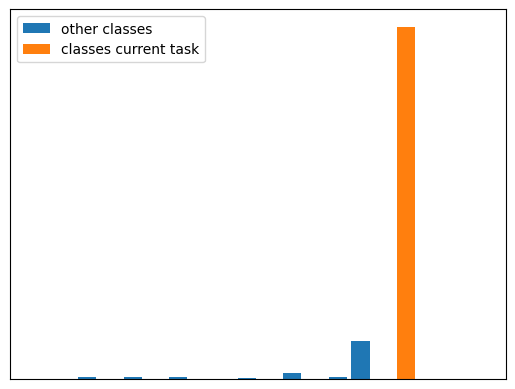}
  \caption{Task 3}
  \label{fig:3}
\end{subfigure}
\end{minipage}\hfil
\begin{minipage}[t]{.35\textwidth}
\begin{subfigure}{\textwidth}
  \includegraphics[width=\linewidth]{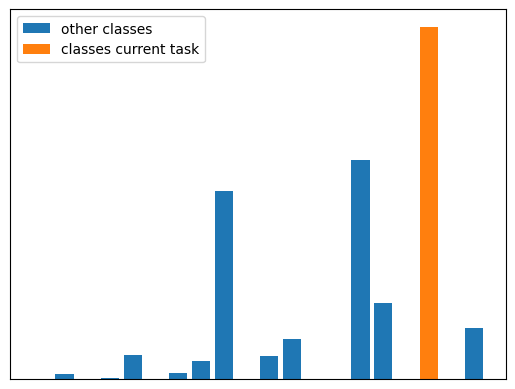}
  \caption{Task 4}
  \label{fig:4}
\end{subfigure}\hfil 
\begin{subfigure}{\textwidth}
  \includegraphics[width=\linewidth]{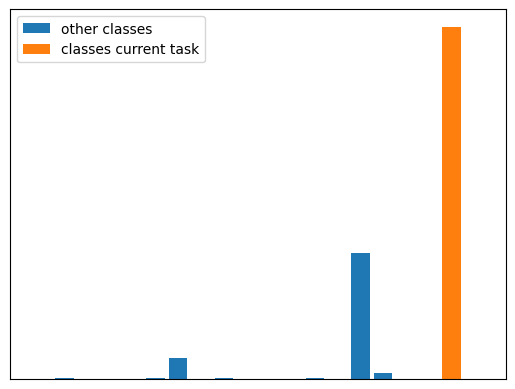}
  \caption{Task 5}
  \label{fig:5}
\end{subfigure}\hfil 
\begin{subfigure}{\textwidth}
  \includegraphics[width=\linewidth]{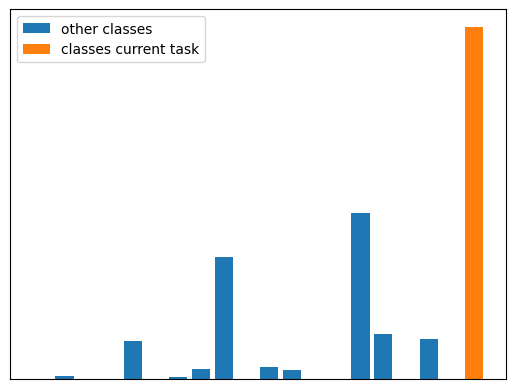}
  \caption{Task 6}
  \label{fig:6}
\end{subfigure}
\end{minipage}
\caption{Classes distribution over tasks for VOC 15p1. In blue the classes without labels.}
\label{fig:freq_15p1}
\end{figure*}

\noindent Figure \ref{fig:freq:10p10} to Figure \ref{fig:freq_15p1} show the class frequencies for each scenario and for each task. In particular, for a given task $t$, in orange we have the classes for task $t$, while in blue, the classes for either old tasks or new ones. These plots are useful to analyze the possible interference between tasks. Moreover, these plots show why Pseudo-label is performing poorly in the scenarios VOC 19p1 and VOC 15p1. For instance, in the 19p1 scenario, many classes for task 1 are missing in the dataset for task 2, namely Pseudo-label is not effective for those classes. This is not the case for replay-based approaches, like RCLPOD. On the other hand, due to the high number of classes, Pseudo-label is more effective in the 40p40 scenario, where instead replay methods struggle due to interference and masking is needed as shown in Sec. \ref{sec:results}.

\subsection{LwF and stability-plasticity dilemma}
\label{ap:lwf}

\noindent As shown in Tab. \ref{tab:lwfvs}, LwF is effective for reducing forgetting but prevents the model from learning a new task. Moreover, in contrast to Pseudo-Label, it prevents forgetting even in the VOC19p1 scenario, where, as shown in the previous section, objects for previous tasks may not appear in the dataset of the new task. These two results are justified by the noisy output of the teacher: contrary to inference, all those predictions for which the model is highly uncertain (very low classification score for any class) are used as target for the student. Therefore, independently by the choice of $\lambda$, LwF constraints the student to mimic the teacher regression output even when it is not necessary.

\subsection{Memory efficiency}
\label{subsec:mem}
In this section,  we provide additional results for the replay-based approaches by changing the memory size to evaluate their impact on performance. In particular, we compare the two best replay-based methods, RCLPOD and OCDM, in the VOC15p1 scenario, where replay-based methods are effective.
From Tab. \ref{tab:mem-size}, we can observe that by decreasing the memory size $m$  from $m=800$ (5\%) to $m=400$ (2.5\%), RCLPOD is less sensitive to this hyper-parameter than OCDM, namely the performance decay slower w.r.t. a memory size decrease.

\subsection{Comparison with literature}

\noindent In Table \ref{tab:litresults}, we report the results obtained in other CLOD works in the 40p40 COCO scenario. Notice that all of them are specifically designed for different architectures with different loss functions, namely Table \ref{tab:litresults} doesn't aim to compare the CL methods but to state the difference of our work based on a more recent Object Detector as YOLOv8. Moreover, we report both the results for YOLOv8n and YOLOv8m, where the latter is comparable in terms of the number of parameters to ResNet50, used as the backbone in the other works.

\subsection{Optimizing Class Distribution in Memory (OCDM)}
\label{subsec:ocdm_pseudocode}
Optimizing Class Distribution in Memory (OCDM) is a greedy approach that selects a subset of samples such that the final distribution of the labels in memory is as close as possible to a uniform target distribution.
OCDM formulates the memory update mechanism as an optimization problem. This greedy algorithm is detailed in Alg. 1.

In particular, once a new batch of data $\mathcal{B}_t$ of size $b_t$ for task $t \in \mathcal{T}$ arrives, the algorithm updates the memory $\mathcal{M}$ of size $M$  by solving the following optimization problem:
\begin{equation}
\begin{aligned}
    \min_{\Omega} \quad & d(\bold{p},\bold{p}_{\Omega}) \\
    \text{subject to} \quad &  \Omega \subseteq \mathcal{M}\cup \mathcal{B}_t \\
                             & |\Omega| = M
\end{aligned}
\end{equation}

where \textbf{p} represents the target distribution i.e. the ideal optimal solution, while, $\textbf{p}_\Omega$ represents the distribution of the labels produced from the samples of the set $\Omega$.
The $d(\cdot,\cdot)$ function used to measure the difference between the two distributions is the Kullback–Leibler (KL) divergence.
The target distribution proposed in \cite{liang2022optimizing} is defined as follows:
\begin{equation}\label{eq:target_distro}
p_i =  \frac{(n_i)^\rho}{\sum_{j = 1}^{C} (n_j)^\rho}  
\end{equation}
where $n_i$ is the frequency of a class $i$ and $\rho$ is the allocation power. Using $\rho = 0$ the samples are saved in memory $\mathcal{M}$ in order to have equally distributed classes.

\bigskip 
\begin{table*}[!htb]
\centering
  \begin{tabular}{lcccc|cccc}
    \toprule
    \multirow{2}{*}{CL Method} &
      \multicolumn{4}{c}{YOLOv8n}&\multicolumn{4}{c}{YOLOv8m}\\
      &{Task 1}& {old} & {new} & {all}&{Task 1}& {old} & {new} & {all}\\
      \midrule
    Pseudo-label &\multirow{2}{*}{34}& 22.7 & 26.0 & 24.4 &\multirow{2}{*}{47.1}&39.2&37.5&38.4 \\
    RCLPOD && 25.7 & 22.5 & 24.1 &&40.9&36.4&38.6 \\
    \bottomrule
  \end{tabular}
  \caption{Results for YOLOv8n and YOLOv8m in the COCO 40p40 scenario. "old" is the mAP for the classes of Task 1 after Task 2 training, "new" is the mAP for the new classes, while "all" is the mAP over all the 80 classes.}
  \label{tab:yolov8m}
\end{table*}
\begin{table*}[!htb]
\centering
  \begin{tabular}{lccc|ccc|ccc}
    \toprule
    \multirow{2}{*}{CL Scenario} &
      \multicolumn{3}{c}{LwF}&\multicolumn{3}{c}{RCLPOD}&\multicolumn{3}{c}{Pseudo-Label}\\
      & {old} & {new} & {all}& {old} & {new} & {all}& {old} & {new} & {all}\\
          VOC10p10 & 50.7 & 63.8 & 57.3 &{\bf 70.9}& 74.1&{\bf 72.5}&68.5&{\bf 74.3}& 71.4\\
    VOC19p1 & 60.6 & 50.3 & 60.0 &{\bf68.8}&61.8& {\bf68.4}&45.8&{\bf 61.9}& 46.6\\
    VOC15p5 & 61.5 & 43.4 & 57.0 &{\bf 71.6}&{\bf 55.8}& {\bf 67.7}&61.1&55.7& 59.7\\
    COCO40p40 & {\bf 26.9} & 11.0 & 19.0 &25.7&22.5& 24.1&22.7&{\bf 26.0}& {\bf24.3}\\
    \bottomrule
  \end{tabular}
  \caption{Comparison LwF, RCLPOD and Pseudo-Label for 2-tasks scenarios. "old" is the mAP for the classes of Task 1 after Task 2 training, "new" is the mAP for the new classes, while "all" is the mAP over all the classes.}
  \label{tab:lwfvs}
\end{table*}

\begin{table*}[!htb]
\centering
  \begin{tabular}{lccc}
    \toprule  
    CL Method &
      $m=800$ & $m=400$& $\Delta\%\;(\downarrow)$\\
      \toprule
    RCLPOD& 56.1&52.5&{\bf 6.4\%}\\
    OCDM & 49.6&43.4&12.5\%\\
    \bottomrule
  \end{tabular}
  \caption{Results for the best replay-based methods in the VOC15p1 scenario varying the memory size $m$.  }
  \label{tab:mem-size}
\end{table*}

\begin{table*}[!htb]
\centering
  \begin{tabular}{lcc}
    \toprule
    
    CL Method &
      mAP 50-95 & Model size\\
      \toprule
    ILOD \cite{shmelkov2017incremental}& 21.30&$\ge 25$M\\
    IncDet \cite{liu2020incdet} & 29.70&$\ge 25$M\\
    Faster ILOD \cite{peng2020faster}& 20.64&$\ge 25$M\\
    SID \cite{peng2021sid}& 25.20&$\ge 25$M\\
    Meta-ILOD \cite{joseph2021incremental}& 23.80&$\ge 25$M\\
    ABR \cite{liu2023augmented}&34.5&$\ge 25$M\\
    RCLPOD (YOLOv8n)&24.09&$3.2$M\\
    RCLPOD (YOLOv8m)&\textbf{38.64}&$25.9$M\\
    \bottomrule
  \end{tabular}
  \caption{MS COCO Incremental 40p40 literature results.}
  \label{tab:litresults}
\end{table*}

\begin{figure*}[ht!]
  \centering
  \begin{subfigure}[b]{0.49\linewidth}
    \centering
    \includegraphics[width=\linewidth]{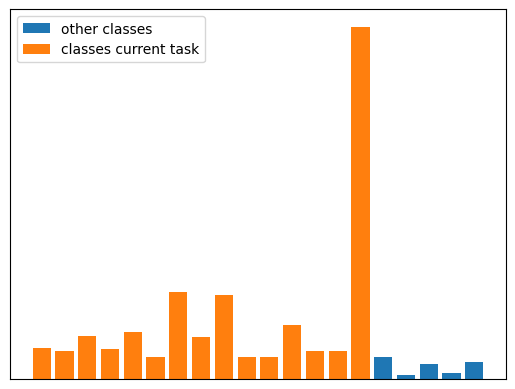}
    \caption{Task 1}
        \label{fig:stats10p10-1}
  \end{subfigure}
    \hfill
  \begin{subfigure}[b]{0.49\linewidth}
    \centering
    \includegraphics[width=\linewidth]{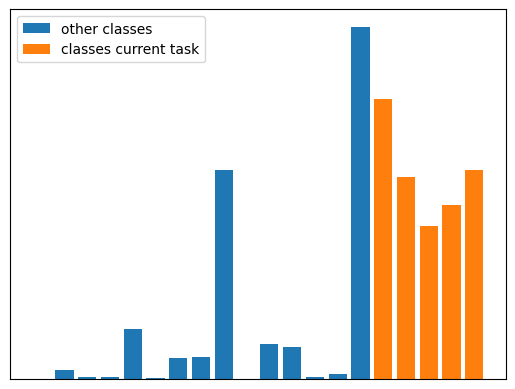}
    \caption{Task 2}
    \label{fig:stats10p10-0}
  \end{subfigure}
  \caption{Classes distribution over tasks for VOC 15p5. In blue the classes without labels.
  }
  \label{fig:plot_results}
\end{figure*}

\clearpage

\begin{algorithm*}
\label{alg:ocdm}
\SetAlgoLined
\DontPrintSemicolon
\SetKwInOut{Input}{Input}
\caption{Optimizing Class Distribution in Memory (OCDM)}
\Input{task stream $\mathcal{T}$, total size of memory M }
\BlankLine
$\mathcal{M} \gets \{\}$ \Comment*[r]{Initialize the memory} 
\For{$t \in \mathcal{T}$} { 
    $\mathcal{D}_t \gets$ Get Dataset $\mathcal{D}_t = \{X_t, Y_t\}$ of task $t$\\

    \BlankLine
    \For{$B_t \in \mathcal{D}_t$}
    {
        \If{$|\mathcal{M}|\leq M$}
        {
            $\text{diff} \gets |\mathcal{M}|-M$\\
            $V_t \gets$ select randomly $\min(|B_t|,\text{diff})$ samples  from $B_t$\\
            $B_t \gets B_t \setminus V_t $\\
            $\mathcal{M} \gets \mathcal{M} \cup V_t$
        }
        \BlankLine
        \If{$|B_t|>0$}
        {
            $\Omega \gets \mathcal{M} \cup B_t$\\
            \For {$k \in [1,2,...,|B_t|]$} {
                 $i = arg \min_{j \in \Omega} d(\bold{p},\bold{p}_{\Omega \setminus \{j\} })$ 
                 
                 $\Omega \gets \Omega \setminus \{i\}$ 
            }
            $\mathcal{M} \gets \Omega$ 
        
    }
    
}
}
\end{algorithm*}

\end{document}